\documentclass{article}

\usepackage[preprint]{neurips_2025}

\usepackage[utf8]{inputenc} 
\usepackage[T1]{fontenc}    
\usepackage{hyperref}       
\usepackage{url}            
\usepackage{booktabs}       
\usepackage{amsfonts}       
\usepackage{nicefrac}       
\usepackage{microtype}      
\usepackage{xcolor}         

\usepackage{microtype}
\usepackage{graphicx}
\usepackage{booktabs} 

\usepackage{multirow}
\usepackage{verbatim}
\usepackage{graphicx}
\usepackage{pifont}
\usepackage{adjustbox}
\usepackage{enumitem}
\usepackage{amsmath,amsfonts}
\usepackage{algorithmic}
\usepackage{algorithm}
\usepackage{array}
\usepackage{caption}

\usepackage{amsmath}
\usepackage{amssymb}
\usepackage{mathtools}
\usepackage{amsthm}
\usepackage{graphicx}
\usepackage{subcaption}
\usepackage{geometry} 
\title{LLaVA-NeuMT: Selective Layer-Neuron Modulation for Efficient Multilingual Multimodal Translation}

\author{
    Jingxuan Wei$^{1}$, Caijun Jia$^{1}\thanks{Equal contribution}$, Qi Chen$^{1*}$, Yujun Cai$^{2}$\thanks{Corresponding author}, Linzhuang Sun$^{1}$, \\
    \textbf{Xiangxiang Zhang$^{1}$, Gaowei Wu$^{1}$, Bihui Yu$^{1}$}  \\
    $^1$University of Chinese Academy of Sciences \\
    $^2$ The University of Queensland  \\
    {\tt\small weijingxuan20@mails.ucas.edu.cn, vanora.caiyj@gmail.com}
}

\begin{document}

\maketitle

\begin{abstract}
Multimodal Machine Translation (MMT) enhances translation quality by incorporating visual context, helping to resolve textual ambiguities. While existing MMT methods perform well in bilingual settings, extending them to multilingual translation remains challenging due to cross-lingual interference and ineffective parameter-sharing strategies. To address this, we propose LLaVA-NeuMT, a novel multimodal multilingual translation framework that explicitly models language-specific and language-agnostic representations to mitigate multilingual interference. Our approach consists of a layer selection mechanism that identifies the most informative layers for different language pairs and a neuron-level adaptation strategy that dynamically selects language-specific and agnostic neurons to improve translation quality while reducing redundancy. We conduct extensive experiments on the M3-Multi30K and M3-AmbigCaps datasets, demonstrating that LLaVA-NeuMT, while fine-tuning only 40\% of the model parameters, surpasses full fine-tuning approaches and ultimately achieves SOTA results on both datasets. Our analysis further provides insights into the importance of selected layers and neurons in multimodal multilingual adaptation, offering an efficient and scalable solution to cross-lingual adaptation in multimodal translation.
\end{abstract}

\section{Introduction}
\begin{figure}[t]
    \centering
    \includegraphics[width=0.8\linewidth]{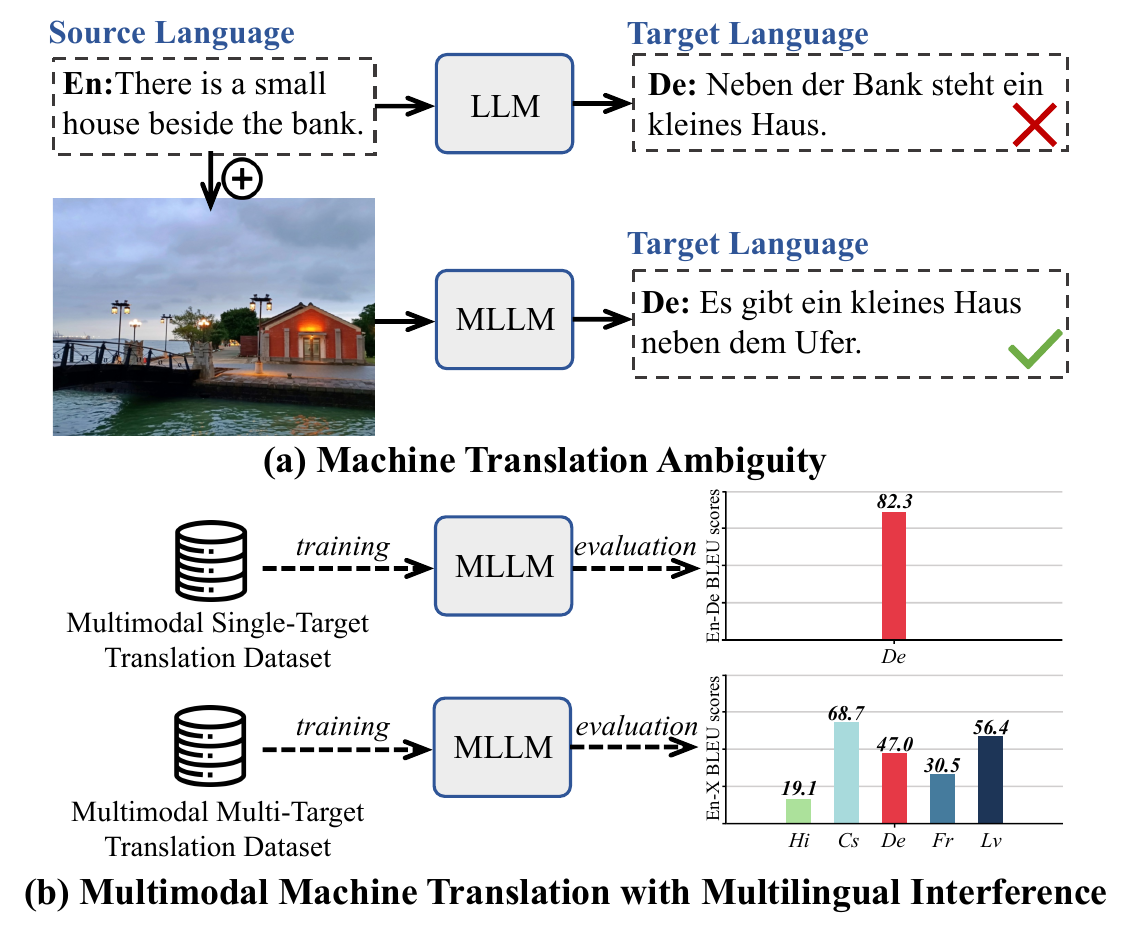}
    \caption{Challenges in MMT.}
    \vspace{-5mm}
    \label{fig:introduction}
\end{figure}

Machine translation has become increasingly crucial in our interconnected world, yet achieving accurate translations remains challenging due to the inherent ambiguities in natural language~\citep{dabre2020survey,klouchek2024bulgarian}. A single word or phrase often carries multiple potential meanings, making it difficult for translation systems to select the appropriate interpretation without additional context. Multimodal Machine Translation (MMT) addresses this challenge by incorporating visual information alongside textual input, helping to resolve ambiguities and improve translation accuracy~\citep{Chen2020, Ma2022, tayir2024encoder}. For example, as shown in Figure~\ref{fig:introduction} (a), when translating the English sentence "There is a small house beside the bank" into German, purely text-based systems often misinterpret "bank" as a financial institution, producing "Neben der Bank steht ein kleines Haus." However, with access to the corresponding image, the system correctly recognizes "bank" as a riverbank and generates the accurate translation "Es gibt ein kleines Haus neben dem Ufer."

While MMT has demonstrated promising results in bilingual settings through various techniques such as multi-task learning, knowledge distillation, and attention mechanisms, extending these approaches to multilingual scenarios presents significant challenges\citep{fan2021beyond,wang2024multimodal}. Multilingual Neural Machine Translation (MNMT) has made progress in text-only translation by leveraging cross-lingual parameter sharing, evolving from simple parameter sharing to more sophisticated approaches like adaptive scheduling and language-specific modules\citep{Jean2019, pan2021contrastive, feng2023towards}. Recently, Mixture-of-Experts (MoE) models have attempted to dynamically allocate computational resources across languages, but often struggle with overfitting and inefficient parameter utilization\citep{fedus2022switch, li2023mmnmt}. Despite these advances, existing MNMT methods exclusively focus on text-based translation and do not address the unique complexities introduced by multimodal information.

As illustrated in Figure~\ref{fig:introduction} (b), multimodal translation in multilingual settings introduces additional challenges beyond those found in either bilingual MMT or text-only MNMT. Recent studies have highlighted that indiscriminate parameter sharing in MNMT can lead to interference between languages, where high-resource languages dominate and degrade the performance of low-resource languages~\citep{shaham2023causes, li2023mmnmt,chen2024pareto}. Furthermore, empirical analysis reveals that different layers in neural translation models serve distinct functions - lower layers often capture general linguistic patterns shared across languages, while higher layers learn language-specific and task-specific features~\citep{tan2024neuron,zhu2024multilingual}. This layered hierarchy becomes particularly crucial in multilingual settings, as different language pairs may rely more heavily on certain layers for effective translation. However, current approaches treat all layers equally when sharing parameters across languages~\citep{Ma2023b, Lan2023, Tian2023}, potentially leading to sub-optimal use of model capacity and increased interference. These observations raise critical questions: How can we identify and leverage the most relevant layers for each language pair? How should we balance parameter sharing across different layers to minimize interference while maintaining translation quality?

To address these challenges, we propose LLaVA-NeuMT, a framework designed to systematically identify and optimize the most relevant model components for each language pair. Our key insight is that selective parameter sharing at both the layer and neuron levels is crucial for balancing effective knowledge transfer and interference mitigation. Instead of sharing all parameters across languages indiscriminately, our method selectively determines which parts of the model are critical for each language pair. First, we introduce a layer selection mechanism that identifies the most informative layers for different language pairs, allowing the model to retain essential representations while reducing computational redundancy. Second, we propose a neuron-level adaptation strategy, where neurons within the selected layers are categorized as either language-specific or language-agnostic based on their activation and gradient variance. Finally, we design a training framework that selectively updates neurons based on the input language pair, mitigating inter-language interference while maintaining computational efficiency.

To validate our approach, we conduct extensive experiments on the M3-Multi30K~\citep{guo2022lvp} and M3-AmbigCaps~\cite{li2021vision} datasets. The results show that LLaVA-NeuMT, utilizing only 40\% of the model parameters, surpasses full fine-tuning baselines. By selecting key layers and fine-tuning language-specific and agnostic neurons, our approach achieves more effective multilingual adaptation. Furthermore, we visualize the importance of selected layers and neurons across languages, offering insights into the adaptation of multimodal translation models.

Our key contributions are as follows:
\begin{itemize}
    \item We propose \textbf{LLaVA-NeuMT}, a multimodal multilingual translation framework that explicitly models \textit{language-specific} and \textit{language-agnostic} representations to mitigate cross-lingual interference in multimodal translation.
    \item We introduce a \textbf{layer and neuron selection mechanism} that identifies the most informative layers and neurons for each language pair, effectively preserving critical representations while reducing redundancy.
    \item We achieve \textbf{SOTA translation performance} across multiple language pairs while fine-tuning a subset of model parameters. Additionally, our analysis provides insights into the importance of different layers and neurons in multimodal multilingual adaptation.
\end{itemize}

\section{Related Work}
\paragraph{Multimodal Machine Translation}
Multimodal Machine Translation (MMT) enhances translation quality by integrating visual context to resolve linguistic ambiguities. Prior research has explored four primary approaches: multi-task learning, knowledge distillation, contrastive learning, and attention-based mechanisms. Multi-task learning integrates OCR and translation models to improve cross-modal representation learning, but these methods often struggle with efficient multilingual adaptation \citep{Chen2020, Ma2022, Su2021b}. To address this, adaptive mechanisms have been introduced to bridge modality gaps and enhance translation consistency \citep{Ma2023b, Lan2023}. Knowledge distillation has been widely used to transfer multimodal knowledge from teacher to student models, ensuring better generalization but often increasing computational overhead \citep{Chen2023, Ma2023c}. Contrastive learning further refines OCR-text alignment and improves robustness in translation tasks, yet remains constrained by reliance on predefined feature mappings \citep{Ma2024, Peng2022a}. Attention-based mechanisms dynamically focus on relevant image regions, improving semantic grounding, but they lack efficient parameter selection for multilingual translation \citep{Mansimov2020, Hinami2021, Jain2021, Tian2023}. While these methods enhance machine translation performance, they often overlook computational efficiency in large-scale multimodal models. As computational demands grow with model size and multilingual adaptation, recent works have emphasized the need to balance model capacity with efficiency \citep{liu2022coupleface,Ma2023a}. However, existing approaches lack fine-grained control over language-specific and agnostic parameters. We address this with a layer-aware neuron modulation framework that improves efficiency and optimizes parameter use.

\paragraph{Multilingual Neural Machine Translation}
Multilingual Neural Machine Translation (MNMT) enables translation across multiple languages within a single model but faces challenges such as inter-language interference and capacity bottlenecks \citep{Aharoni2019, fan2021beyond,ijcai2024p722}. Prior works address these issues through adaptive scheduling \citep{Jean2019,pan2021contrastive}, gradient-based optimization \citep{wang2020balancing, feng2023towards}, and language-specific modules \citep{philip2020monolingual, zhang2021share}. Mixture-of-Experts (MoE) models allocate capacity dynamically \citep{fedus2022switch,li2023mmnmt}, though overfitting remains a concern. Recent studies highlight that indiscriminate parameter sharing degrades high-resource language performance \citep{huang2024survey,nimma2024comparative, javed2025transformer}, leading to strategies such as binary masks \citep{poppi2024towards} and contrastive learning \citep{liang2024continual} to mitigate interference. 
However, these approaches often introduce additional complexity and computational costs. While research on multilingual interference has primarily focused on text-based models \citep{Jean2019,li2023mmnmt, javed2025transformer}, its implications for multimodal translation remain insufficiently studied. 
In contrast, we introduce a selective layer and neuron-level modulation framework to optimize multilingual adaptation, reducing interference while maintaining efficiency in multimodal MNMT.

\section{Methodology}

\begin{figure*}[t]
    \centering
    \includegraphics[width=\linewidth]{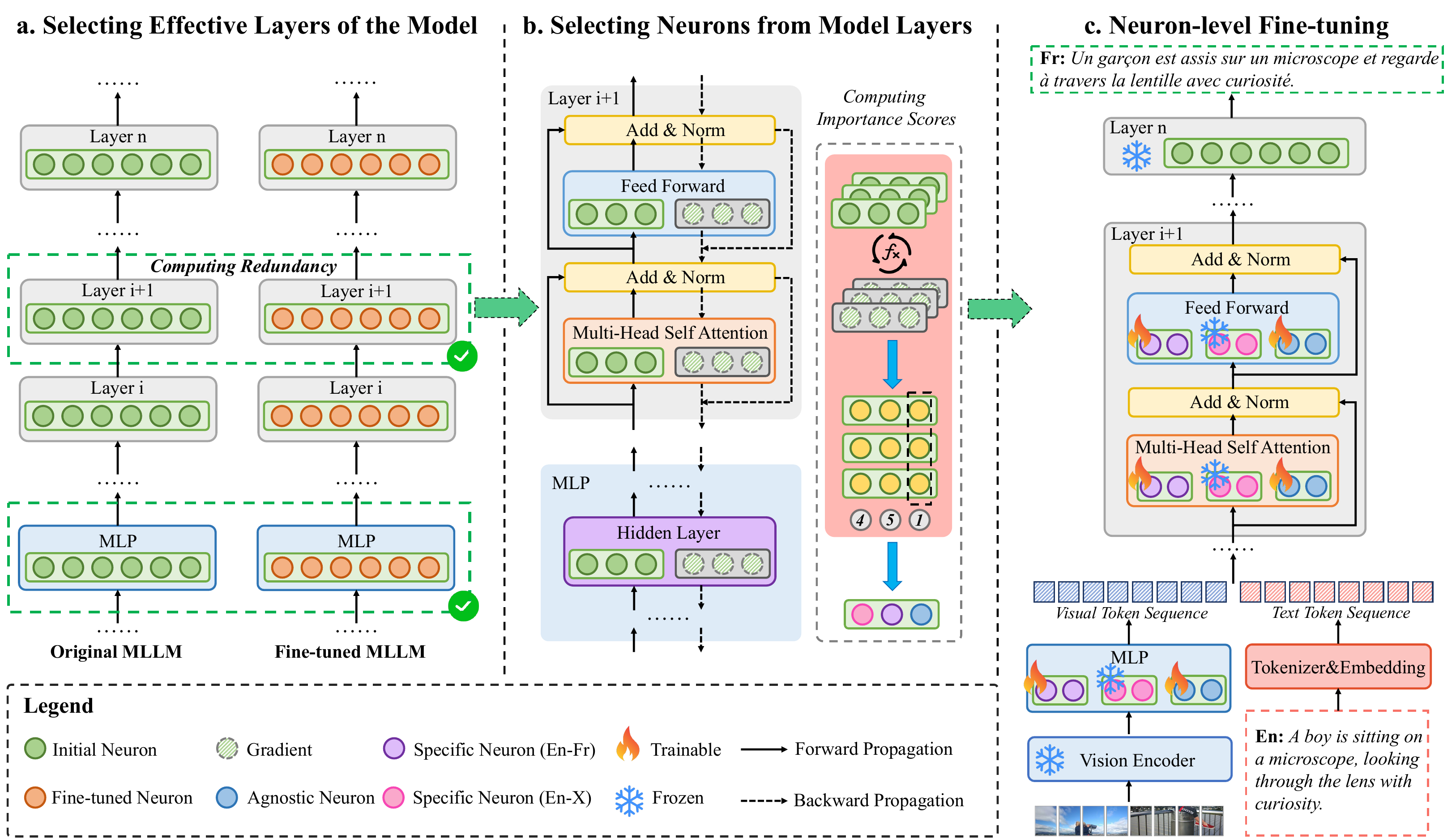}
    \caption{LLaVA-NeuMT Model Architecture.}
    \vspace{-3mm}
    \label{fig:layer_selection}
\end{figure*}

\subsection{Multimodal Machine Translation}
Multimodal machine translation extends traditional machine translation by incorporating visual information to enhance contextual understanding. Given a source sentence $X^s$ in language $s$, a corresponding image $I$, and a target language $t$, the objective is to generate a translated sentence $Y^t$ that preserves the semantics of the source sentence while leveraging visual context. The translation process can be formulated as a function $\mathcal{F}$ that maps the source text and image to the target text:

\begin{equation}
    Y^t = \mathcal{F}(X^s, I, s, t; \theta),
\end{equation}

where $\theta$ represents the model parameters. The model encodes textual features through a text encoder $\mathcal{E}_t$ and extracts visual features using a vision encoder $\mathcal{E}_v$:

\begin{equation}
    \mathcal{T} = \mathcal{E}_t(X^s), \quad \mathcal{V} = \mathcal{E}_v(I).
    \label{eq:multimodal_encoding}
\end{equation}

The extracted textual and visual features are combined within a multimodal translation model, producing an intermediate representation that is subsequently decoded into the target language.

\subsection{Selecting Effective Layers of the Model}
\label{sec:layer_selection}
In MMT, different layers of the model contribute differently to text and image processing. To improve efficiency while maintaining translation quality, and inspired by \citep{li2024adaptive,wangcluster}, we introduce a layer selection method that identifies and retains the most informative layers in both the vision-language connector and the large language model (LLM). Given a model with $L$ layers, the objective is to determine a subset $\mathcal{L} \subseteq \{1,2,\dots,L\}$ that maximizes task relevance while reducing redundancy.

The importance of each layer is assessed based on activation similarity before and after supervised fine-tuning (SFT), as illustrated in Figure~\ref{fig:layer_selection} (a). Given activations from layer $l$ in the pretrained model ($X^A_l$) and the fine-tuned model ($X^B_l$), the redundancy-based importance score $R_l$ is computed as:

\begin{equation}
    R_l = \frac{(X^A_l \cdot X^B_l)^2}{\|X^A_l\|^2 \|X^B_l\|^2 + \epsilon},
\end{equation}

where $X^A_l$ and $X^B_l$ are the activations from layer $l$ in the pretrained and fine-tuned models, and $\epsilon$ is a small constant to avoid numerical instability. A lower $R_l$ value suggests that a layer undergoes significant adaptation during fine-tuning, indicating its importance for the translation task. Layers are ranked based on $R_l$, and a subset $\mathcal{L}_s$ is selected corresponding to the top $\alpha$ fraction of layers, where $\alpha$ is a tunable hyperparameter.

\subsection{Selecting Important Neurons of the Model}
\label{sec:neuron_selection}
In MMT, different neurons within each selected layer contribute differently to various language pairs \citep{zhu2024landermt}. To further optimize the model, we introduce a neuron selection mechanism that identifies the most relevant neurons for each language pair while preserving generalizable neurons across all languages. Given a layer $l$ with $N$ neurons, our objective is to classify neurons into two categories: \textit{language-specific neurons} and \textit{language-agnostic neurons}.

Neuron selection is performed on the previously selected layers, as illustrated in Figure~\ref{fig:layer_selection} (b). The importance of each neuron $n$ is evaluated based on its activation and gradient values during fine-tuning. Specifically, for each training instance, we compute the importance score as:

\begin{equation}
    \mathcal{I}_n = | A_n \times G_n |
\end{equation}

where $A_n$ and $G_n$ represent the activation and backpropagation gradient of neuron $n$, respectively. Given $K$ language pairs, we aggregate the importance scores over $T$ training samples and compute the variance across languages as:

\begin{equation}
    \sigma^2(n) = \frac{1}{K} \sum_{k=1}^{K} \left( \mathcal{I}_n^k - \bar{\mathcal{I}}_n \right)^2
\end{equation}

where $\mathcal{I}_n^k$ denotes the importance score of neuron $n$ for language pair $k$, and $\bar{\mathcal{I}}_n$ represents the mean importance score across all language pairs.

To classify neurons, we first define \(\mathcal{N}\) as the set of all neurons in the selected layers:


\begin{equation}
    \mathcal{N} = \{ n \mid n \in \mathcal{N}_l, l \in \mathcal{L}_s \}
\end{equation}

where \(\mathcal{L}_s\) is the selected layer set and \(\mathcal{N}_l\) the neurons in layer \(l\). We then define two subsets:

\begin{equation}
    \mathcal{S}_k = \{ n \in \mathcal{N} \mid \mathcal{I}_n^k = \max_{j} \mathcal{I}_n^j \}
\end{equation}

where $\mathcal{S}_k$ represents the set of language-specific neurons, which exhibit the highest importance for a single language pair $k$.

\begin{equation}
    \mathcal{A} = \{ n \in \mathcal{N} \mid \sigma^2(n) \leq \epsilon \}
\end{equation}

where $\mathcal{A}$ represents the set of language-agnostic neurons, which maintain relatively stable importance scores across all language pairs, with a variance threshold $\epsilon$.

\subsection{LLaVA-NeuMT}
LLaVA-NeuMT performs neuron-level adaptation based on the previously selected layers and classified neurons, as illustrated in Figure~\ref{fig:layer_selection} (c). Given a source sentence $X^s$ in language $s$, an image $I$, and a target language $t$, the model extracts features using a text encoder $\mathcal{E}_t$ and a vision encoder $\mathcal{E}_v$, producing textual and visual representations as defined in Equation~(\ref{eq:multimodal_encoding}). These representations are passed through the selected layers $\mathcal{L}_s$ (Section~\ref{sec:layer_selection}), where neuron updates are applied selectively. Based on the classification of neurons into language-specific ($\mathcal{S}_k$) and language-agnostic ($\mathcal{A}$) categories (Section~\ref{sec:neuron_selection}), only relevant neurons receive parameter updates during fine-tuning.

To achieve this, a gradient masking mechanism is applied to constrain updates to neurons belonging to $\mathcal{S}_k$ or $\mathcal{A}$. Specifically, for each neuron $n$, the gradient is modified as follows:

\begin{equation}
    G'_{n} =
    \begin{cases}
    G_n, & \text{if } n \in \mathcal{A} \cup \mathcal{S}_k \\
    0, & \text{otherwise}
    \end{cases}
\end{equation}

where $G_n$ represents the computed gradient of neuron $n$. Neurons outside these sets are frozen, preventing unnecessary parameter updates.

The final model update is performed using:

\begin{equation}
    \theta_n \leftarrow \theta_n - \eta G'_n
\end{equation}

where $\eta$ is the learning rate. By restricting updates to selected neurons, LLaVA-NeuMT efficiently adapts the model while maintaining stability across different language pairs. This fine-tuning strategy ensures that multimodal representations are effectively adapted, allowing both textual and visual features to be optimized for MNMT.

\section{Experiments}
\subsection{Experimental Setting}
\paragraph{Datasets} 
We evaluate our approach on two multimodal MNMT datasets: M3-Multi30K~\citep{guo2022lvp} and M3-AmbigCaps~\cite{li2021vision}. M3-Multi30K consists of 29,000 image-text translation pairs for training and 1,000 for testing, covering multiple language pairs. M3-AmbigCaps is a larger dataset with 89,600 training pairs and 1,000 test pairs, designed for evaluating multimodal translation performance.

\paragraph{Experimental Setup} 
We adopt LLaVA-1.5-7B~\citep{liu2024improved} as the pretrained backbone and optimize training using DeepSpeed ZeRO-3 on 4 × A100 (80GB) GPUs. The model is trained for 4 epochs with a per-device batch size of 16 and a gradient accumulation step of 1. Mixed precision training with BF16 is applied to reduce memory overhead. The optimizer is AdamW with a learning rate of 2e-5 and a cosine annealing scheduler, with 3\% warmup. Weight decay is set to 0. The maximum text sequence length is 2048, and image inputs are resized to a fixed aspect ratio, with visual features extracted from the second-to-last layer of the Vision Transformer (ViT)~\citep{nguyen2024image}.

\paragraph{Evaluation Metrics \& Baselines} 
We evaluate translation performance using BLEU-4. Baselines include Text-only MT models: Text Transformer~\citep{fan2021beyond}; Open-source MMT models: Qwen2-VL~\citep{wang2024qwen2}, MiniCPM~\citep{yao2024minicpm}, InternVL~\citep{chen2024internvl}; Closed-source MMT models: GPT-4o~\citep{achiam2023gpt}, Gemini-1.5-Pro~\citep{team2024gemini}; and Multimodal MNMT models: Vision Matters (Gated Fusion)~\citep{li2021vision}, Vision Matters (Concatenation)~\citep{li2021vision}, LVP-M3~\citep{guo2022lvp}, and the multilingual fine-tuned version of LLaVA-1.5~\citep{liu2024improved}.

\subsection{Main Results and Analysis}

\begin{table*}[t]
  \centering
      \caption{BLEU scores on the M3-Multi30K test set. \textbf{Best} results are bold, \underline{second-best} underlined.}
  \resizebox{\linewidth}{!}{
    \begin{tabular}{lllllllll}
    \toprule
    Type  & Model (En→X) & Fr    & Cs    & De    & Lv    & Hi    & Tr    & Avg-{all} \\
    \midrule
    Text-only MT  & Text Transformer~\citep{fan2021beyond} & 61.8  & 32.8  & 40.6  & 51.2  & 59.0    & 53.8  & 49.8 \\
    \midrule
    \multirow{3}[0]{*}{Open-source MMT} & Qwen2-VL-7B~\citep{wang2024qwen2} & 44.1  & 7.8   & 33.5  & 0.1   & 0.6   & 0.8   & 14.5 \\
          & MiniCPM-2.6-8b~\citep{yao2024minicpm} & 26.2  & 4.0     & 27.2  & 0.1  & 0.2   & 0.3   & 9.7 \\
          & InternVL-2.5-7b~\citep{chen2024internvl} & 35.2  & 8.2   & 25.8  & 0.1   & 3.3   & 1.0     & 12.3 \\
          \midrule
    \multirow{2}[0]{*}{Closed-source MMT} & GPT-4o~\citep{achiam2023gpt} & 53.8  & \textbf{37.4}  & \textbf{44.3}  & 39.4  & 28.3  & 28.6  & 38.6 \\
          & Gemini-1.5-Pro~\citep{team2024gemini} & 38.5  & 22.2  & 23.5  & 24.3  & 10.4  & 22.4  & 23.6 \\
          \midrule
    \multirow{4}[0]{*}{Multimodal MNMT} & Vision Matters (Gated fusion)~\citep{li2021vision} & 62.5  & 32.9  & 41.2  & 52.1  & 59.6  & 54.2  & 50.4 \\
          & Vision Matters (Concatenation)~\citep{li2021vision}  & 59.7  & 33.1  & 39.8  & 50.3  & 57.6  & 51.4  & 48.6 \\
          & LVP-M3~\citep{guo2022lvp} & 63.7 & 34.6 & \underline{43.2} & 53.5 & 61.4 & 55.6 & 52.0 \\
          & LLaVA-1.5-SFT(default)~\citep{liu2024improved} & 66.5 & 35.9 & 42.2 & 56.1 & \underline{61.5} & 57.8 & 53.3 \\
          \midrule
    \multirow{2}[0]{*}{Ours}   &LLaVA-NeuMT (40\%) & \textbf{67.0}    & \underline{36.0}    & 42.0    & \underline{57.3}  & 60.0    & \underline{58.3}  & \underline{53.4} \\
      & LLaVA-NeuMT (80\%)  & \underline{66.8} & 35.9 & 42.6 & \textbf{58.2} & \textbf{61.8} & \textbf{60.7} & \textbf{54.3} \\
    \bottomrule
    \end{tabular}
    }
  \label{tab:m3_multi30k}
\end{table*}

\begin{table*}[t]
  \centering
      \caption{BLEU scores on the M3-AmbigCaps test set. \textbf{Best} results are bold, \underline{second-best} underlined.}
  \resizebox{\linewidth}{!}{
    \begin{tabular}{lllllllll}
    \toprule
    Type  & Model (En→X) & Fr    & Cs    & De    & Lv    & Hi    & Tr    & Avg\_all \\
    \midrule
    Text-only MT  & Text Transformer~\citep{fan2021beyond} & 62.3  & 47.8  & 49.0    & 46.6  & 52.4  & 35.9  & 49.0 \\
    \midrule
    \multirow{3}[2]{*}{Open-source MMT} & Qwen2-VL-7B~\citep{wang2024qwen2} & 40.3  & 2.7   & 27.3  & 0.3   & 0.6   & 0.7   & 12.0 \\
          & MiniCPM-2.6-8b~\citep{yao2024minicpm} & 32.6  & 2.8   & 19.8  & 0.1   & 0.15  & 0.2   & 9.3 \\
          & InternVL-2.5-7b~\citep{chen2024internvl}  & 31.6  & 6.16  & 10.7  & 0.1   & 3.3   & 0.8   & 8.8 \\
    \midrule
    \multirow{2}[2]{*}{Closed-source MMT} & GPT-4o~\citep{achiam2023gpt} & 43.6  & 29.6  & 38.0    & 26.5  & 24.9  & 16.7  & 29.9 \\
          & Gemini-1.5-Pro~\citep{team2024gemini} & 28.8  & 13.6  & 18.3  & 15.9  & 12.2  & 12.2  & 16.8 \\
    \midrule
    \multirow{4}[2]{*}{Multimodal MNMT} & Vision Matters (Gated fusion)~\citep{li2021vision} & 64.3  & 50.3  & 51.2  & 48.5  & 54.1  & 38.7  & 51.2 \\
          & Vision Matters (Concatenation)~\citep{li2021vision}   & 62.6  & 47.6  & 48.7  & 45.9  & 52.7  & 36.0    & 48.9 \\
          & LVP-M3~\citep{guo2022lvp} & 65.7  & 52.9  & 53.7  & 51.6  & 56.3  & 42.7  & 53.8 \\
          & LLaVA-1.5-SFT(default)~\citep{liu2024improved}  & 72.1  & \underline{57.3}  & 60.3  & \underline{56.5}  & \underline{56.8}  & 45.2  & 58.0 \\
    \midrule
    \multirow{2}[0]{*}{Ours}  &LLaVA-NeuMT (40\%) & \underline{73.2}  & 57.0  & \underline{60.9}  & 56.2   & 56.5    & \underline{46.2}  & \underline{58.3} \\
      & LLaVA-NeuMT (80\%)  & \textbf{74.1} & \textbf{58.4} & \textbf{61.7} & \textbf{57.8} & \textbf{58.4} & \textbf{47.9} & \textbf{59.7} \\
    \bottomrule
    \end{tabular}
    }
  \label{tab:m3_ambigcaps}
\end{table*}%

We evaluate the performance of different models across four categories, as shown in Table~\ref{tab:m3_multi30k} and Table~\ref{tab:m3_ambigcaps}, Text-only MT achieves strong results, demonstrating that textual models alone can provide high-quality translations. However, it still underperforms compared to Multimodal MNMT, which integrates visual context to improve translation quality. Open-source MMT models show significantly lower performance, particularly in low-resource languages such as Latvian, Hindi, and Turkish, likely due to the lack of multimodal multilingual training data, which limits their generalization in multilingual settings. Closed-source MMT models, such as GPT-4o, achieve competitive results in high-resource languages but show a noticeable drop in low-resource scenarios, suggesting that general-purpose multimodal models are not optimized for multilingual translation. In contrast, Multimodal MNMT models consistently achieve better BLEU scores, confirming that incorporating multimodal signals benefits multilingual translation. Among them, LLaVA-1.5-SFT enhances translation quality through supervised fine-tuning. Our proposed LLaVA-NeuMT further improves performance while fine-tuning only 40\% of the model parameters, demonstrating the efficiency of selective layer adaptation. When increasing the fine-tuned layers to 80\%, LLaVA-NeuMT achieves the best results, showing that balancing layer selection and neuron modulation enhances translation performance while maintaining efficiency. Additionally, our fine-tuning strategy, which adjusts language-specific and agnostic neurons at a 1:9 ratio, ensures effective multilingual adaptation. In terms of language-specific trends, GPT-4o performs well on high-resource languages such as French, Czech, and German in the M3-Multi30K test set but struggles in lower-resource languages. The performance gap is more evident in the M3-AmbigCaps test set, where the larger dataset scale and increased task complexity further challenge general-purpose models. By contrast, LLaVA-NeuMT consistently outperforms other models across both datasets, demonstrating its robustness in Multimodal MNMT.

\subsection{Effect of Layer Selection on MMT}
\label{Effect_of_layer}
\begin{figure}[t]
    \centering
    \begin{subfigure}[b]{0.48\linewidth}
        \includegraphics[width=\linewidth]{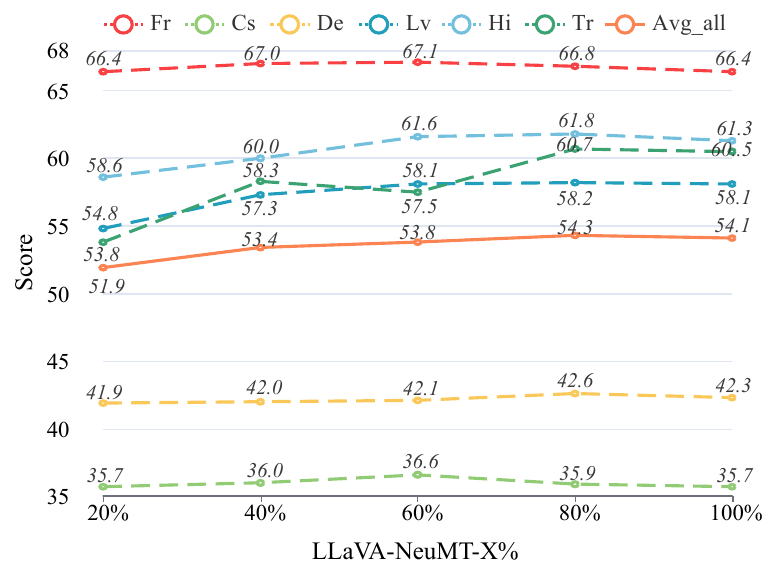}
        \caption{Effect of layer selection on translation. x\% indicates the top x\% most important layers.}
        \label{fig:layer_impact}
    \end{subfigure}
    \hfill
    \begin{subfigure}[b]{0.48\linewidth}
        \includegraphics[width=\linewidth]{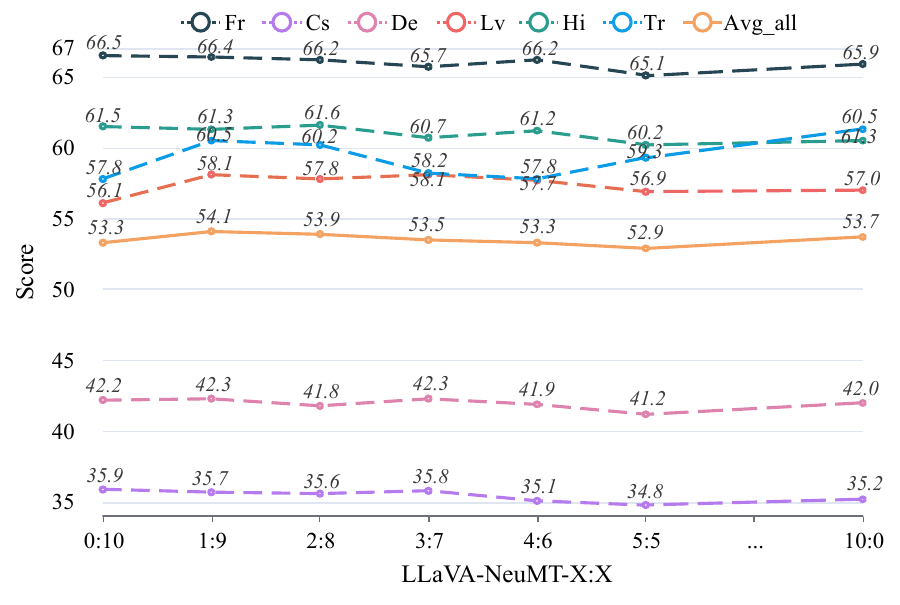}
        \caption{Impact of specific-to-agnostic neuron ratio on translation performance on the M3-Multi30K dataset.}
        \label{fig:neuron_ratio_impact}
    \end{subfigure}
    \caption{Translation performance under different model configurations.}
    \label{fig:layer_and_neuron_impact}
\end{figure}

To investigate the role of layer selection in multimodal multilingual translation, we evaluate performance by selecting the top 20\%, 40\%, 60\%, 80\%, and 100\% most important layers, ranked by importance scores computed in Section~\ref{sec:layer_selection}. In this experiment, the neuron selection strategy remains fixed, with language-specific and agnostic neurons adjusted at a 1:9 ratio, ensuring that the only variable is the number of selected layers. As shown in Figure~\ref{fig:layer_impact}, BLEU scores increase as more layers are included, reaching the highest performance at 80\% selection. Beyond this point, performance declines, suggesting that retaining all layers introduces redundancy or noise, negatively impacting translation quality. Using only 20\% of the layers leads to significantly lower BLEU scores, indicating that a minimal subset is insufficient for effective multimodal multilingual adaptation. Between 40\% and 80\%, all language pairs exhibit consistent improvements, with the most pronounced gains observed in low-resource languages such as Latvian, Hindi, and Turkish. For high-resource languages such as French, Czech, and German, performance stabilizes beyond 60\% and slightly decreases at 100\%, reinforcing that excessive layers do not necessarily contribute positively to translation. These findings demonstrate that an optimal layer selection strategy enhances translation quality while maintaining efficiency, with 80\% selection striking the best balance between performance and computational cost.

\subsection{Effect of Neuron Selection on MMT}
\begin{table}[t]
    \centering
    \caption{Effect of agnostic and specific neurons on multimodal multilingual translation on the M3-Multi30K dataset. "Standard" denotes a 1:9 specific-to-agnostic neuron ratio, while "Agnostic" and "Specific" refer to models fine-tuning only agnostic or specific neurons.}
    \resizebox{\linewidth}{!}{
    \begin{tabular}{lccccccc}
    \toprule
    En→X & Fr & Cs & De & Lv & Hi & Tr & Avg\_all \\
    \midrule
    Standard & 66.4 & 35.7 & 42.3 & 58.1 & 61.3 & 60.5 & 54.1 \\
    Agnostic & 65.8 & 35.6 & 41.2 & 56.3 & 61.1 & 59.7 & 53.3 \\
    Specific & 65.9 & 35.2 & 42.0 & 57.0 & 60.5 & 61.3 & 53.7 \\
    \bottomrule
    \end{tabular}
    }
    
    \label{tab:neuron_impact}
\end{table}

To analyze the impact of agnostic and specific neurons in multimodal multilingual translation, we conduct experiments where all layers are selected while varying the neurons that are fine-tuned. As shown in Table~\ref{tab:neuron_impact}, the highest BLEU score is achieved when both neuron types are optimized in a 1:9 ratio. Fine-tuning only agnostic neurons results in a slight performance drop, while fine-tuning only specific neurons leads to a further decline. This suggests that while specific neurons contribute to translation quality, agnostic neurons play a more crucial role in ensuring multilingual adaptation. The relatively competitive performance of fine-tuning specific neurons alone indicates that language-specific features remain valuable, particularly in distinguishing linguistic variations. However, the performance gap between agnostic-only and specific-only settings reinforces the greater importance of agnostic neurons in maintaining stable multilingual translation. Examining language-specific trends, Czech benefits more from fine-tuning agnostic neurons, suggesting a stronger dependence on cross-lingual representations, whereas Turkish achieves its highest accuracy when only specific neurons are fine-tuned, indicating that some languages rely more on task-specific adaptation.

To further examine the effect of adjusting the ratio of specific to agnostic neurons, we conduct experiments while keeping all layers selected. As shown in Figure~\ref{fig:neuron_ratio_impact}, increasing the proportion of specific neurons initially improves BLEU scores, peaking at a 1:9 ratio. Beyond this point, performance declines as the proportion of agnostic neurons decreases, suggesting that excessive specific neurons may reduce generalization ability. However, at extreme ratios (e.g., 10:0), performance slightly rebounds, indicating that in certain cases, heavily relying on specific neurons can still capture relevant translation patterns. This suggests that while an optimal balance of neuron types is necessary, models exhibit some degree of robustness when specific neurons dominate. Across different language pairs, Czech exhibits a steady decline when agnostic neurons are reduced, confirming its reliance on agnostic representations. In contrast, Hindi and Turkish maintain relatively stable performance across different neuron ratios, demonstrating adaptability to both neuron types. These findings emphasize the necessity of a well-balanced allocation of agnostic and specific neurons for optimal MMT.

\subsection{Visualization of Layer Importance}

\begin{figure}[t]
    \centering
    \begin{subfigure}[b]{0.48\linewidth}
        \includegraphics[width=\linewidth]{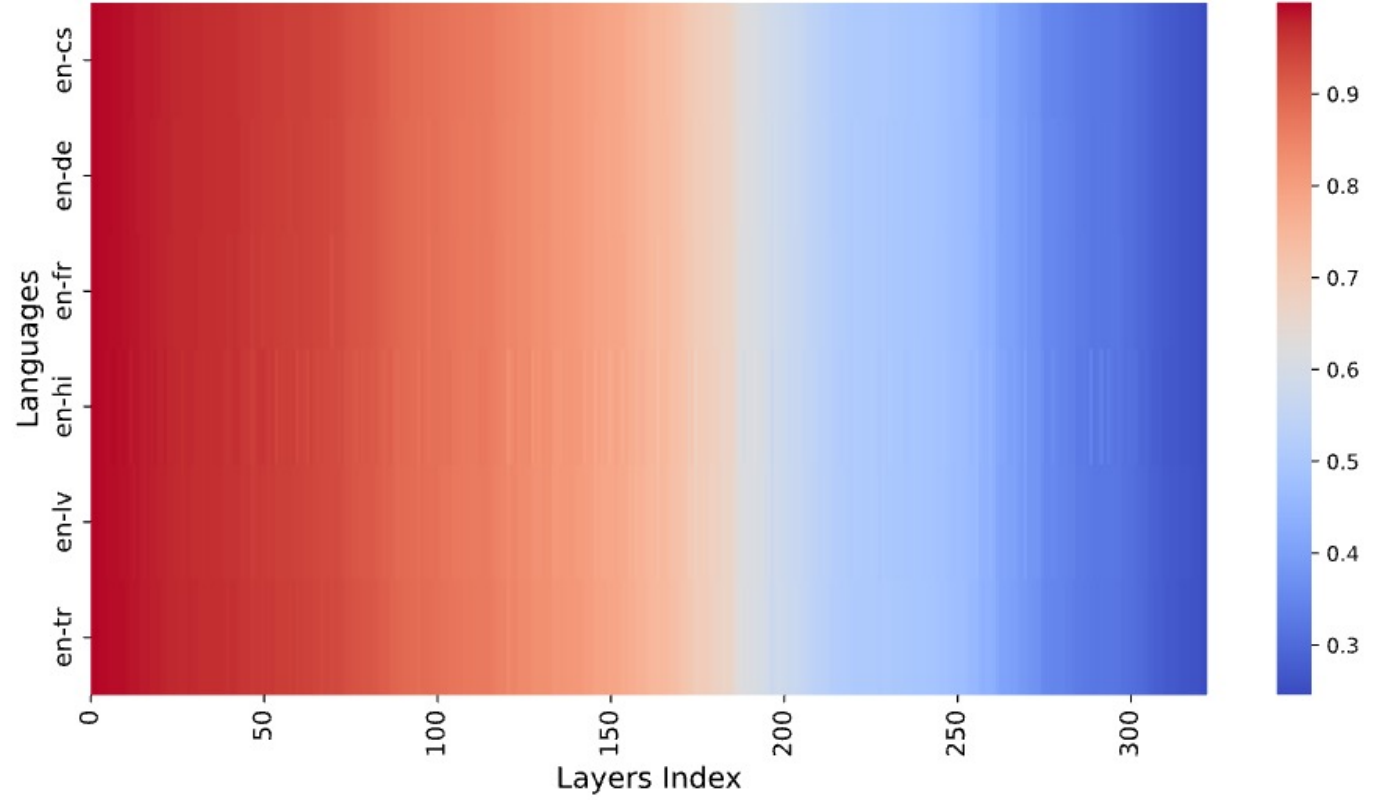}
        \caption{Layer importance across language pairs.}
        \label{fig:layer_importance}
    \end{subfigure}
    \hfill
    \begin{subfigure}[b]{0.48\linewidth}
        \includegraphics[width=\linewidth]{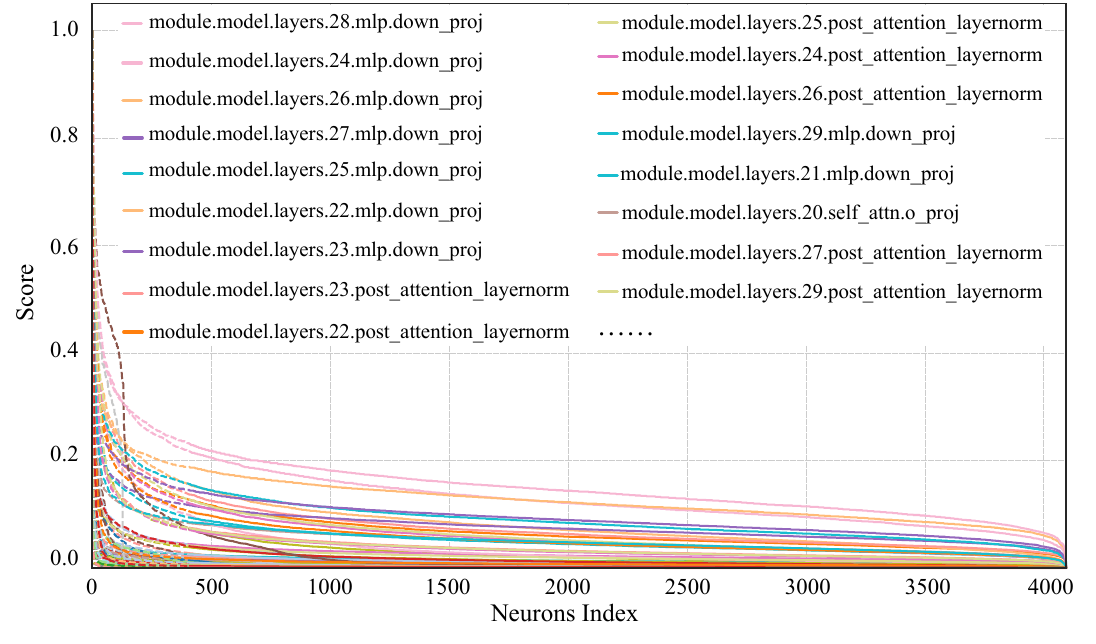}
        \caption{Neuron importance across selected layers.}
        \label{fig:neuron_importance}
    \end{subfigure}
    \vspace{-3mm}
    \caption{Layer and neuron analysis on M3-Multi30K.}
    \label{fig:layer_neuron_viz}
\end{figure}

To investigate the role of different layers in multimodal multilingual translation, we visualize the layer importance scores across multiple language pairs in Figure~\ref{fig:layer_importance}. The x-axis represents model layers, the y-axis denotes language pairs, and the color intensity indicates relative importance. The heatmap reveals significant variations in layer importance across the model, demonstrating that selecting key layers is necessary rather than uniformly fine-tuning all layers. From a horizontal perspective, the first 250 layers (approximately 80\% of the model depth) exhibit relatively high importance scores, consistently exceeding 0.5. This trend aligns with the findings in Section~\ref{Effect_of_layer}, where selecting the top 40-80\% of layers resulted in optimal translation performance. The concentration of importance in these layers suggests that they capture essential multimodal and multilingual representations. From a vertical perspective, the importance scores remain relatively stable across different language pairs, indicating that layer selection is primarily influenced by architectural properties rather than specific language characteristics. This confirms that an effective layer selection can enhance computational efficiency without significantly affecting translation quality across languages.

\subsection{Analysis of Specific and agnostic neurons}
To investigate the distinction between specific and agnostic neurons in multimodal multilingual translation, we visualize neuron importance variance across six language pairs in Figure~\ref{fig:neuron_importance}. We select the top 40\% of layers (108 layers) and observe that in each layer, a small subset of neurons exhibits significantly higher variance, indicating their language specificity. This confirms the necessity of differentiating specific and agnostic neurons rather than treating them uniformly. From a distribution perspective, the first 10\% of neurons in each layer (dashed lines) display high variance, while the remaining 90\% (solid lines) maintain stable scores. This supports our choice of a 1:9 ratio between specific and agnostic neurons, ensuring an optimal balance between language adaptability and cross-lingual generalization. Furthermore, we identify key neuron types critical to multimodal multilingual translation, including attention projection layers and MLP down-projection layers. Unlike conventional large language models, which primarily rely on deep linguistic representations, multimodal translation models emphasize connector layers for effective cross-modal alignment, underscoring their importance in improving translation quality.

\section{Conclusion and Limitation}
In this work, we tackled multilingual interference in MMT by introducing LLaVA-NeuMT, a framework that selectively optimizes layers and neurons to enhance efficiency and translation quality. Our approach integrates a layer selection mechanism to retain the most informative layers and a neuron-level adaptation strategy to balance language-specific and agnostic representations. Experiments on the M3-Multi30K and M3-AmbigCaps datasets show that LLaVA-NeuMT achieves SOTA performance while fine-tuning fewer parameters. Further analysis reveals that selecting 40-80\% of layers yields optimal results, and a 1:9 specific-to-agnostic neuron ratio effectively balances generalization and adaptation. Future work will explore adaptive parameter-sharing strategies and extend our approach to broader multilingual and multimodal scenarios.

While effective, our current design relies on fixed thresholds for layer and neuron selection, which may not generalize across all languages or domains. Future work will explore adaptive strategies and extend our approach to broader multilingual and multimodal settings.


\bibliography{custom}
\bibliographystyle{abbrv}


\appendix



\newpage

\section{Broader Impact}
\vspace{-2mm}

This work addresses the important challenge of multilingual multimodal machine translation (MMMT), where models must handle both linguistic diversity and cross-modal alignment. We present LLaVA-NeuMT, a framework designed to improve translation quality and efficiency across diverse language pairs and visual contexts.

\paragraph{Accessibility and Social Value} Enhancing translation for low-resource languages and varied visual domains can promote more equitable access to information. Our approach has potential applications in education, assistive technologies, multilingual captioning, and cross-cultural communication, particularly for underrepresented communities.

\paragraph{Reliability and Responsible Use} As with other large-scale multimodal models, misuse is possible if the system is deployed without safeguards—for example, generating inaccurate translations or reinforcing spurious image-text alignments. While LLaVA-NeuMT does not introduce new risks beyond typical architectures, we encourage responsible deployment practices, including transparent evaluation and task-specific constraints.

\paragraph{Bias and Representation} The model may reflect biases present in its training data and may perform unevenly across languages. These are known limitations in multilingual NLP and multimodal systems. Addressing such biases and improving representation for under-resourced languages remains an important direction for future work.

\paragraph{Efficiency and Sustainability} By fine-tuning only a selective subset of parameters, our method reduces training cost and energy consumption. This contributes to more sustainable and scalable development of multilingual and multimodal systems.


\end{document}